\documentclass[letterpaper, 10 pt, conference]{ieeeconf}  

\IEEEoverridecommandlockouts                              

\usepackage{graphicx}

\usepackage{amsmath} 
\usepackage{bm}

\usepackage[noend]{algorithmic}
\usepackage{algorithm}

\usepackage{subfig}
\usepackage{array}

\title{\LARGE \bf
Robot Local Planner: A Periodic Sampling-Based Motion Planner with Minimal Waypoints for Home Environments
}

\author{Keisuke Takeshita$^{1}$, Takahiro Yamazaki$^{1}$, Tomohiro Ono$^{1}$ and Takashi Yamamoto$^{2}$
\thanks{*This research was partially supported by grants from The Nitto Foundation and the Ichihara International Scholarship Foundation.}
\thanks{$^{1}$ Keisuke Takeshita, Takahiro Yamazaki and Tomohiro Ono are with Frontier Research Center, Toyota Motor Corporation, Toyota, Aichi, Japan.}%
\thanks{{\tt\small keisuke\_takeshita@mail.toyota.co.jp}}
\thanks{$^{2}$ Takashi Yamamoto is with Department of Information Science, Faculty of Information Science, Aichi Institute of Technology, Toyota, Aichi, Japan.}%
}

\begin{document}

\maketitle
\thispagestyle{empty}
\pagestyle{empty}

\begin{abstract}
The objective of this study is to enable fast and safe manipulation tasks in home environments.
Specifically, we aim to develop a system that can recognize its surroundings and identify target objects while in motion, enabling it to plan and execute actions accordingly.
We propose a periodic sampling-based whole-body trajectory planning method, called the “Robot Local Planner (RLP).”
This method leverages unique features of home environments to enhance computational efficiency, motion optimality, and robustness against recognition and control errors, all while ensuring safety.
The RLP minimizes computation time by planning with minimal waypoints and generating safe trajectories.
Furthermore, overall motion optimality is improved by periodically executing trajectory planning to select more optimal motions.
This approach incorporates inverse kinematics that are robust to base position errors, further enhancing robustness.
Evaluation experiments demonstrated that the RLP outperformed existing methods in terms of motion planning time, motion duration, and robustness, confirming its effectiveness in home environments.
Moreover, application experiments using a tidy-up task achieved high success rates and short operation times, thereby underscoring its practical feasibility.
\end{abstract}


\section{INTRODUCTION}
The global labor shortage due to declining birth rates and aging populations is a critical issue.
Additionally, promoting the independence of people with disabilities and the elderly is also an important challenge.
To address these issues, we are advancing the development of a safe and compact domestic support robot called a Human Support Robot (HSR) \cite{robomechj_hsr}.

One of our primary goals is to enable the rapid and safe execution of manipulation tasks in home environments.
Specifically, we aim to develop a system that can recognize its surroundings and target objects while in motion, allowing it to plan and execute actions accordingly.
To ensure safety, the system incorporates a function that halts operations when a task becomes infeasible.
In this study, we discuss whole-body trajectory planning suitable for this manipulation system.

Whole-body trajectory planning must satisfy the following four key criteria:
\begin{enumerate}
    \item Fast computation --- The system must respond quickly to perception, requiring short computation times.
    \item Motion optimality --- Unnecessary movements should be eliminated to reduce motion duration.
    \item Robustness --- The system must effectively handle perception and control errors, as recognition occurs during motion.
    \item Safety --- Whole-body trajectories must prevent collisions with the environment and avoid self-collision.
\end{enumerate}

Existing methods cannot satisfy all these requirements simultaneously.
Model predictive control \cite{storm}, control-based planning using machine learning \cite{mpinets}\cite{spin}, and motion-optimization-based planning \cite{chomp}\cite{stomp}\cite{trajopt} offer short computation times and high motion optimality; however, they struggle to ensure safety.
Conversely, while sampling-based motion planning methods \cite{rrt}\cite{rrt_connect} guarantee safety, they tend to have long computation times.
Methods that consider motion optimality \cite{aitstar} involve a trade-off between computation time and motion optimality.


In this study, we propose a periodic sampling-based whole-body trajectory planning method, termed the Robot Local Planner (RLP).
The RLP leverages the unique characteristics of home environments to enhance computational efficiency, motion optimality, and robustness while ensuring safety.
In home environments, objects are typically arranged for human convenience, reducing the need for dense trajectory sampling.
Thus, the RLP employs only two types of trajectories:
\begin{enumerate}
    \item Straight-line trajectory: A direct path connecting the current robot state to the goal state.
    \item Three-point trajectory: A trajectory that passes through a randomly selected intermediate robot state.
\end{enumerate}
This approach enables the RLP to efficiently identify collision-free whole-body trajectories while reducing computation time.
Safety is maintained through a sampling-based approach.
Additionally, periodic planning improves motion optimality by allowing transitions to more optimal trajectories.
To further enhance robustness, we incorporate inverse kinematics (IK), which is robust to base position errors.

The contributions of this study are as follows:
\begin{enumerate}
    \item Proposal of a periodic sampling-based whole-body trajectory planning method called the RLP.
    \item Verification that motion planning with minimal waypoints is feasible in home environment settings.
    \item Implementation of a tidy-up task using an actual robot equipped with the RLP.
\end{enumerate}

\begin{figure*}[tb]
    \centering
    \vspace{6pt}
    \includegraphics[width=0.95\hsize]{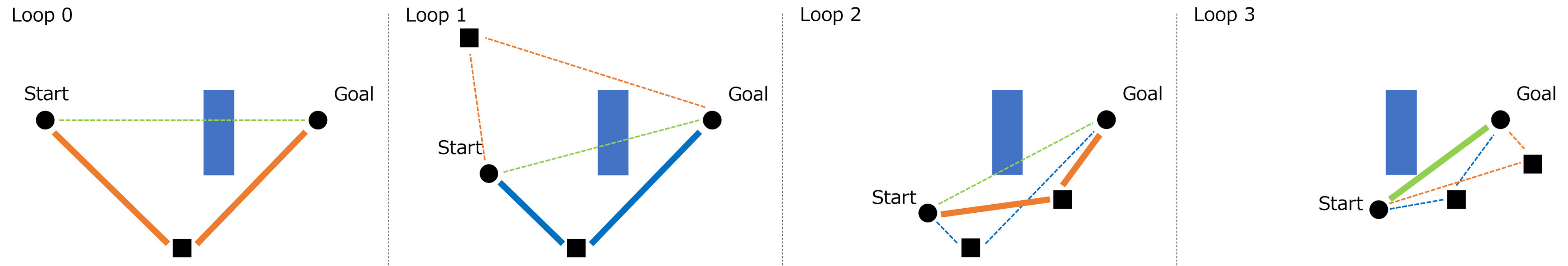}
    \vspace{-6pt}
    \caption{Trajectory planning of the RLP.
             Blue squares represent obstacles.
             Green lines denote straight-line trajectories, orange lines indicate three-point trajectories, and blue lines show previously selected trajectories.
             Solid lines represent the selected trajectories, which are the shortest collision-free trajectories, while dashed lines indicate unselected trajectories.}
    \label{fig:rlp_idea}
\end{figure*}

\section{RELATED WORK}

\subsection{Motion Planning}
Motion planning for a mobile manipulator is a path-planning problem in the joint space, which includes base positions and orientations.
The Rapidly-exploring Random Tree (RRT) \cite{rrt} efficiently handles high-dimensional path planning by expanding a tree from the start through random sampling.
RRT-Connect \cite{rrt_connect} is a representative method that expands the RRT search tree from both the start and goal points.
Although RRT-Connect is among the faster methods in the RRT series, its computation time remains relatively long.
Adaptively Informed Trees (AIT*) \cite{aitstar} utilize heuristics for rapid optimal pathfinding, balancing the trade-off between computation time and motion optimality.

Optimization-based methods, such as CHOMP \cite{chomp}, STOMP \cite{stomp}, and TrajOpt \cite{trajopt}, optimize trajectories using metrics like path smoothness and distance from obstacles; however, they do not guarantee safety.

Motion planning methods also utilize reactive motion control.
For instance, STORM \cite{storm} achieves fast model predictive control by computing collision costs using a Graphics Processing Unit.
Additionally, Motion Policy Networks \cite{mpinets} and SPIN \cite{spin} train models to control robots using obstacle point clouds or depth images.
However, these methods do not guarantee the safety of the generated trajectories.

\subsection{Robustness Metrics}
A key metric for assessing the robustness in robot states is manipulability \cite{manipulability}.
Manipulability indicates the ease of moving the robot’s end-effector, contributing to robustness against changes in target positions.
Vahrenkamp et al. \cite{manipulability_analysis} extended this concept by incorporating factors such as collisions and joint limits; however, challenges remain regarding computation time.

Studies emphasizing robustness against perception and control errors have also been conducted.
Foulon et al. \cite{724648} introduced a metric that penalizes base movement.
Yamazaki et al. \cite{YAMAZAKI2022104232} proposed a method that accounts not only for base movement but also for recognition errors of the grasped object and the gripper’s tolerance. 
While these approaches enhance robustness by considering uncertainties, they also increase computational load.
Conversely, Takeshita et al. \cite{robust_ik_ad_2024} proposed a method to rapidly identify IK solutions that are robust to base movement errors.

\section{PROBLEM SETTING AND APPROACH}

\subsection{Definition of Target Manipulation Task}
In this subsection, we define the target manipulation tasks in a domestic environment.
The robot is guided to the task location by a navigation system.
Upon recognizing the target object, the robot simultaneously executes multiple motion-planning techniques.
The planning space is approximately 2 m wide.
Concurrent planning is advantageous because different methods have varying strengths and weaknesses \cite{orthey2023sampling}, and it is a common practice \cite{moveit}.
Our RLP method aims to achieve fast computation, motion optimality, robustness, and safety in scenarios where ample space is available for both human and robot movements.

\subsection{Approach}
In general, the RLP operates similarly to RRT-Connect in motion, performing planning periodically, as shown in Fig. \ref{fig:rlp_idea}.
\begin{itemize}
    \item Loop 0: The RLP identifies a collision-free three-point trajectory (orange), and the robot follows it.
    \item Loop 1: Both the previously selected trajectory (blue) and a newly sampled three-point trajectory (orange) remain collision-free. The RLP selects the shorter, previously chosen trajectory.
    \item Loop 2: A new, shorter three-point trajectory (orange) is sampled, prompting the robot to transition to this more efficient trajectory.
    \item Loop 3: A straight-line trajectory (green) is identified as collision-free and the shortest option, leading the robot to transition to this trajectory.
\end{itemize}
The RLP efficiently generates safe and highly optimal trajectories with minimal computation time.

\section{METHOD}

\subsection{Robot Local Planner}
\begin{figure}[tb]
    \centering
    \vspace{6pt}
    \includegraphics[width=0.97\hsize]{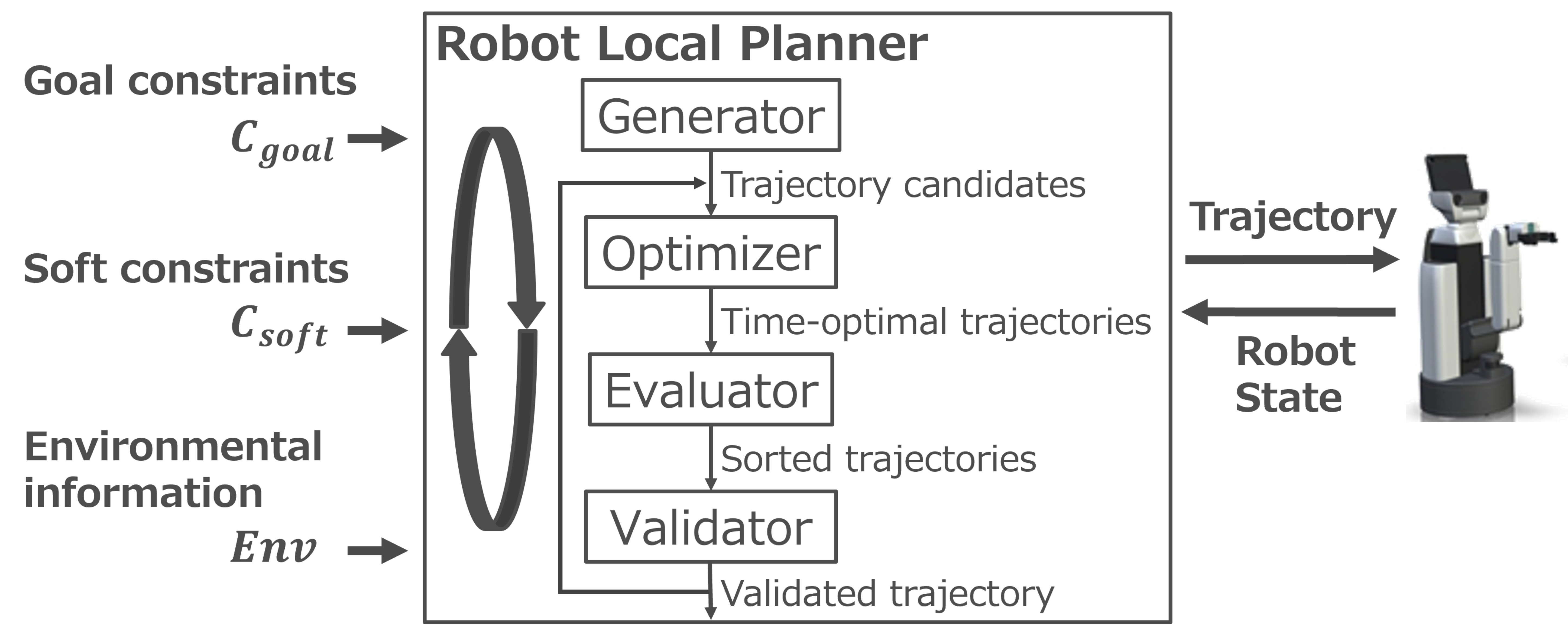}
    \caption{System overview of the RLP.
             The robot's whole-body trajectory is planned through the processes of trajectories generation, time optimization, evaluation, and validation.}
	\vspace{-8pt}
    \label{fig:rlp_system}
\end{figure}

The RLP processing flow is illustrated in Fig. \ref{fig:rlp_system}.
The RLP sequentially executes the Generator, Optimizer, Evaluator, and Validator based on the constraints, environmental information, and robot state.
This process produces a time-optimized whole-body trajectory that satisfies the constraints and prevents collisions by publishing the updated trajectory at periodic intervals $T_p$.

The input information for the RLP is as follows:
\begin{itemize}
    \item Goal constraints $\bm{C}_{goal}$: Conditions that the trajectory must satisfy at its endpoint.
    \item Soft constraints $\bm{C}_{soft}$: Conditions that define the desirable trajectories.
    \item Environmental information $Env$: Obstacle data used for collision detection.
    \item Robot state $\bm{q}_{init}$: Joint state of the robot, encompassing its base x-position, y-position, yaw rotation, and their respective velocities.
\end{itemize}
Each of $\bm{C}_{goal}$ and $\bm{C}_{soft}$ contains either the desired end-effector pose range $C_{ee}$ or the range of robot joint angles/base positions $C_{joint}$.
Task Space Region \cite{tsr} defines $C_{ee}$, whereas $C_{joint}$ is determined by the limits of the base pose and the position of each joint.
$\bm{q}_{init}$ uses the predicted state after $T_p$ based on the current state.

If a trajectory exists during motion, it is added to the set of candidates generated by the Generator.
This enables motion planning to account for the current trajectory.
Specifically, if a more optimal trajectory is generated, the RLP transitions to it; otherwise, the current trajectory is maintained.
Additionally, if the current trajectory becomes unsuitable due to environmental changes, the RLP can select another collision-free trajectory or stop if no viable trajectory is found.
This approach enables adaptation to dynamically changing environments.

\subsection{Generator}
The Generator produces a set of trajectories consisting of straight lines and three-point trajectories that satisfy any of the goal constraints $\bm{C}_{goal}$.

The processing flow is presented in Algorithm \ref{alg:generator_overall}.
Trajectory generation continues until either the generation time exceeds $TO_{gen}$ or the number of generated trajectories reaches $N_Q$. 
$N_s$ denotes the maximum number of straight-line trajectories in the trajectory set.

First, the desired end-effector pose constraints $\bm{C}_{ee}$ is extracted from the goal constraints $\bm{C}_{goal}$.
If $\bm{C}_{ee}$ exists, the SAMPLE\_WITH\_ROBUST\_IK process (Algorithm \ref{alg:generator_with_robust_ik}) is executed to generate a set of trajectories using robust IK solutions.
Next, until either the number of generated trajectories reaches $N_Q$ or the generation time exceeds $TO_{gen}$, random trajectories are generated using SAMPLE\_TRAJECTORY (Algorithm \ref{alg:generator_random}).
The reason for not using only robust IK solutions is to ensure diversity among the trajectories.
Relying solely on robust IK solutions results in low diversity, thereby reducing the probability of generating collision-free trajectories.

Next, we describe the SAMPLE\_WITH\_ROBUST\_IK process (Algorithm \ref{alg:generator_with_robust_ik}), which generates a set of trajectories using robust IK solutions.
CHOICE randomly selects a constraint.
SOLVE\_IK is a function that employs an algorithm \cite{robust_ik_ad_2024} to determine a set of IK solutions robust to base position control errors.
Robustness is defined as "the expected value of the existence of continuous IK solutions that can achieve the same desired end-effector pose even when base position control errors occur" and is mathematically expressed as:
\begin{equation}
  R(\bm{q}(x_0, y_0)) = \int f(\bm{q}(x_0, y_0), x, y) p(x, y | x_0, y_0) dx dy.  \label{eq:robustness_con}
\end{equation}
Here, $\bm{q}(x_0, y_0)$ represents the whole-body IK solution at the base position $(x_0, y_0)$.
$f(\bm{q}(x_0, y_0), x, y)$ is a function that returns 1 if there exists a continuous whole-body IK solution at the base position $(x, y)$ that can achieve the same end-effector pose as $\bm{q}(x_0, y_0)$ and 0 otherwise.
$p(x_0, y_0, x, y)$ is the probability distribution of moving to the base position $(x, y)$ when controlling for the base position $(x_0, y_0)$.

\begin{figure}[t]
    \centering
    \vspace{-6pt} 
\end{figure}
\begin{algorithm}[t]
    \caption{Processing flow of the Generator}
    \label{alg:generator_overall}
    \begin{algorithmic}[1]
        \REQUIRE $\bm{C}_{goal}, \bm{q}_{init}, Env$
        \STATE $trajectories$ = []
        \STATE $\bm{C}_{ee}$ = EXTRACT($\bm{C}_{goal}$)
        \IF {NOT $\bm{C}_{ee}$.EMPTY()}
            \STATE $trajectories$ = SAMPLE\_WITH\_ROBUST\_IK(\\
                \hspace{12pt}$\bm{C}_{ee}, \bm{q}_{init}, Env$)
        \ENDIF
        \STATE $use\_middle$ = false
        \FOR {$i = trajectories$.SIZE() ... $N_Q - 1$}
            \IF {IS\_TIMEOUT($TO_{gen}$)}
                \STATE Break
            \ENDIF
            \IF {$i \ge N_s$}
                \STATE $use\_middle$ = true
            \ENDIF
            \STATE $trajectories$.APPEND(SAMPLE\_TRAJECTORY(\\
                \hspace{12pt}$\bm{C}_{goal}, \bm{q}_{init}, use\_middle$))
        \ENDFOR
        \STATE Return $trajectories$
    \end{algorithmic}
\end{algorithm}
\begin{figure}[t]
    \centering
    \vspace{-20pt} 
\end{figure}

Depending on the number of generated trajectories, either a straight-line or a three-point trajectory is generated.
SAMPLE\_RANDOM outputs a random robot state.
The base pose is derived by adding a uniform random number with width $R_p$ for the position and $R_r$ for the rotation to the midpoint between $\bm{q}_{init}$ and $\bm{q}_{goal}$.
In addition, each joint takes a random value within its limits.
MERGE is a function that combines the input trajectory points and outputs them as trajectories.

Furthermore, we describe the SAMPLE\_TRAJECTORY process (Algorithm \ref{alg:generator_random}), which generates random trajectories.
CHOICE randomly selects the constraint $C_{goal}$.
Then, SAMPLE\_FROM\_CONSTRAINT generates a random desired state $\bm{q}_{goal}$ that satisfies $C_{goal}$.
The subsequent process is identical to that described in Algorithm \ref{alg:generator_with_robust_ik}.

\subsection{Optimizer}
The Optimizer performs time optimization on a set of trajectories by incorporating the current velocity.
This process utilizes an extended version of the method proposed by Kunz et al. \cite{Kunz2012TimeOptimalTG}, which accounts for the current velocity of a robot.
The resulting time-optimized trajectories are sampled at time intervals of ${T}_{s}$.

\begin{figure}[t]
    \centering
    \vspace{-6pt} 
\end{figure}
\begin{algorithm}[t]
    \caption{SAMPLE\_WITH\_ROBUST\_IK}
    \label{alg:generator_with_robust_ik}
    \begin{algorithmic}[1]
        \REQUIRE $\bm{C}_{ee}, \bm{q}_{init}, Env$
        \STATE $C_{ee}$ = CHOICE($\bm{C}_{ee}$)
        \STATE List[$\bm{q}_{goal}]$ = SOLVE\_IK($C_{ee}, Env$)
        \STATE $trajectories$ = []
        \FOR{$\bm{q}_{goal}$ in List[$\bm{q}_{goal}$]}
            \IF{$trajectories$.SIZE() $< N_s$}
                \STATE $candidate$ = MERGE($\bm{q}_{init}$, $\bm{q}_{goal}$)
            \ELSE
                \STATE $\bm{q}_{middle}$ = SAMPLE\_RANDOM($\bm{q}_{init}$, $\bm{q}_{goal}$)
                \STATE $candidate$ = MERGE($\bm{q}_{init}$, $\bm{q}_{middle}$, $\bm{q}_{goal}$)
            \ENDIF
            \STATE $trajectories$.APPEND($candidate$)
        \ENDFOR
        \STATE Return $trajectories$
    \end{algorithmic}
\end{algorithm}
\begin{algorithm}[t]
    \caption{SAMPLE\_TRAJECTORY}
    \label{alg:generator_random}
    \begin{algorithmic}[1]
        \REQUIRE $\bm{C}_{goal}, \bm{q}_{init}, {use}\_{middle}$
        \STATE $C_{goal}$ = CHOICE($\bm{C}_{goal}$)
        \STATE $\bm{q}_{goal}$ = SAMPLE\_FROM\_CONSTRAINT($C_{goal}$)
        \IF{${use}\_{middle}$}
            \STATE $\bm{q}_{middle}$ = SAMPLE\_RANDOM($\bm{q}_{init}$, $\bm{q}_{goal}$)
            \STATE Return MERGE($\bm{q}_{init}$, $\bm{q}_{middle}$, $\bm{q}_{goal}$)
        \ELSE
            \STATE Return MERGE($\bm{q}_{init}$, $\bm{q}_{goal}$)
        \ENDIF
    \end{algorithmic}
\end{algorithm}

\subsection{Evaluator}
The Evaluator ranks the time-optimal trajectories using soft constraints $\bm{C}_{soft}$, sorting the trajectories in descending order of evaluation values.

Using the trajectory $\bm{Q} = \{ (\bm{q}_0, t_0), ..., (\bm{q}_n, t_n) \}$ and soft constraints $\bm{C}_{soft}$, the evaluation value $e(\bm{Q}, \bm{C}_{soft})$ is calculated as follows:
\begin{equation}
  e(\bm{Q}, \bm{C}_{soft}) = - t_n + \sum\nolimits_{C \in \bm{C}_{soft}} e(C, \bm{Q}).
\end{equation}
Here, $C$ represents a constraint and $e(C, \bm{Q})$ is a function that returns the evaluation value of the trajectory $\bm{Q}$ under constraint $C$.
The function is designed so that higher evaluation values for each constraint, along with shorter motion durations, contribute to a greater overall evaluation value.

\subsection{Validator}
The Validator checks the trajectories for the collisions using the environmental information.
The checks continue until one of the following conditions is satisfied: the verification time exceeds $TO_{val}$, a collision-free trajectory is identified, or all trajectories have been checked.
If a collision-free trajectory is found, it is selected; otherwise, the process fails.

The trajectory validation process is outlined in Algorithm \ref{alg:validator}. 
First, EXTRACT\_POINTS extracts the indices of the trajectory points from trajectory $\bm{Q}$ used for collision checking.
To accelerate periodic planning, collision checks are dense for recent points and sparse for others.
All trajectory points from the current to ${VT}_{from\_start}$ are checked.
Beyond this range, collision checks are conducted at intervals of $VT_{interval}$.
SAMPLE\_ENV derives obstacle information $collisions$ for each robot state $\bm{q}_i$ at time $t_i$, requiring $\bm{q}_i$ to reflect grasped objects.
Finally, IS\_COLLISION verifies collisions based on $\bm{q}_i$ and $collisions$.
If all trajectory points are collision-free, the process is successful.

\begin{figure}[t]
    \centering
    \vspace{-6pt} 
\end{figure}
\begin{algorithm}[t]
    \caption{Process of validating a trajectory}
    \label{alg:validator}
    \begin{algorithmic}[1]
        \REQUIRE $\bm{Q}, Env$
        \STATE $indices$ = EXTRACT\_POINTS($\bm{Q}$)
        \FOR{$i$ in $indices$}
            \STATE $collisions$ = SAMPLE\_ENV($Env, \bm{q}_i, t_i$)
            \IF{IS\_COLLISION($\bm{q}_i, collisions$)}
                \STATE Return FAILURE
            \ENDIF
        \ENDFOR
        \STATE Return SUCCESS
    \end{algorithmic}
\end{algorithm}

\section{EXPERIMENTS}

\subsection{System for Experiments}
\begin{figure}[ht]
    \centering
    \includegraphics[width=85mm]{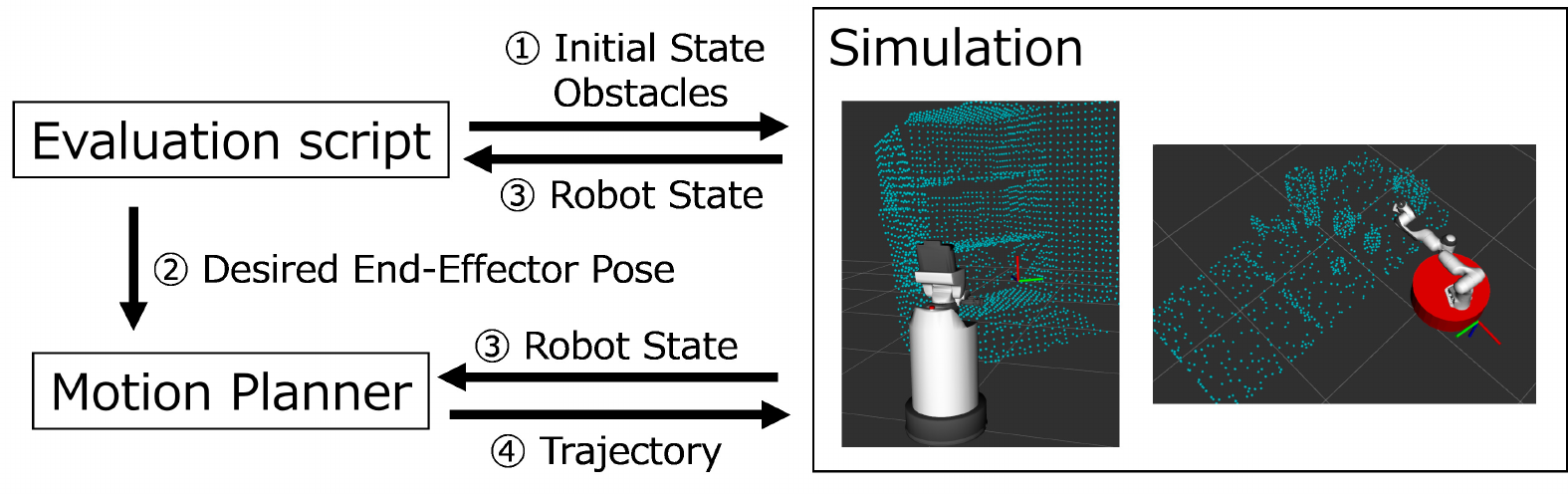}
    \caption{Experimental system}
	\vspace{-8pt}
    \label{fig:eval_sys}
\end{figure}

The experimental system setup is shown in Fig. \ref{fig:eval_sys}.
In each experiment, the evaluation script read the desired end-effector pose, obstacle point cloud, and initial robot state. 
The script initialized the robot state and obstacle point clouds in the simulation, then commanded the trajectory planning to achieve the desired end-effector pose.
The trajectory planning generated and sent trajectories to the simulator based on the command values and the robot state of the simulator. 
Simultaneously, the script collected the robot state from the simulator.
The experiment ended when the end-effector position error was \textless 0.01 m and the rotational error was \textless 15 deg, or after 20 s.

\begin{figure*}[tb]
    \centering
    \vspace{6pt}
    \includegraphics[width=0.97\hsize]{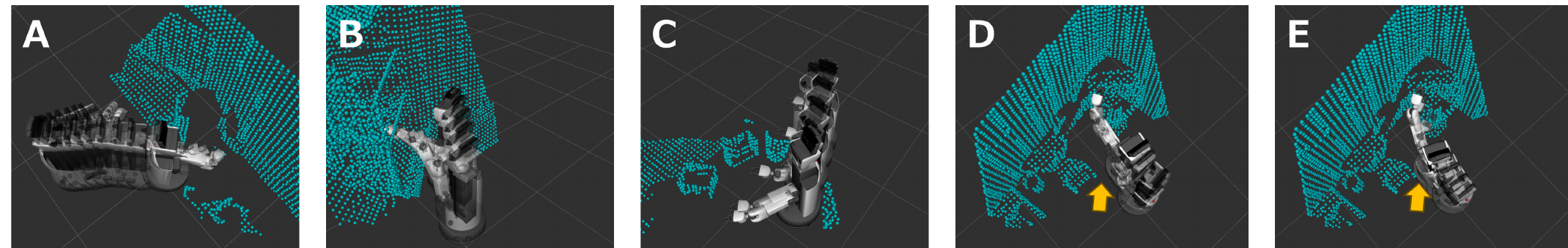}
    \vspace{-4pt}
    \caption{Examples of robot trajectories and their environments are shown. 
             D shows an example when robust IK is utilized, showing a tendency to maintain distance from obstacles to avoid collisions due to base position control errors.
             In contrast, E shows when robust IK is not used.
             In this case, the robot is closer to obstacles compared to D, resulting in lower robustness.}
    \label{fig:trajectory}

\end{figure*}

\subsection{Performance Evaluation in Home Environments}
The performance of the proposed method was evaluated in home environments using multiple robot models and various evaluation input sets designed for these settings.

The robot models used in the experiment included the HSR (featuring three and five degrees of freedom (DOF) for the base and arm, respectively) \cite{robomechj_hsr} and a mobile manipulator equipped with a 7-DOF Franka Emika Panda mounted on an omnidirectional base (Franka MM). 
The methods compared were the proposed RLP, AIT*, and CBiRRT2 \cite{tsr} (an extension of RRT-Connect).
We used data from the BEHAVIOR Dataset \cite{behavior_dataset} for the evaluation, which is used for imitation learning in home environments \cite{robust_ik_ad_2024}.
We only included the evaluation inputs for which a solution could be obtained using CBiRRT2, thereby excluding the evaluation inputs for which trajectory generation was impossible for each robot.

For the RLP implementation, the following parameters were set: $T_p$ to 0.25 s, $N_Q$ to 50, $N_s$ to 5, $R_p$ to 2.0 m, and $R_r$ to 1.5 rad.
For HSR, SAMPLE\_FROM\_CONSTRAINT in Algorithm \ref{alg:generator_random} uses a Hybrid IK \cite{robomechj_hsr}, whereas Franka MM employs a numerical IK \cite{numeric_ik_sugihara}.
The initial base position for IK was randomly set to within 2 m in the x- and y-directions from the desired end-effector position, and the initial orientation was random, with the joint angles set randomly within their limits.
Considering the difference in computation speed of IK, the $TO_{gen}$ was set to 0.1 s for the HSR and 0.5 s for the Franka MM, respectively.
In SOLVE\_IK of Algorithm \ref{alg:generator_with_robust_ik}, "robust IK solutions" are defined as those with at least 0.5 × maximum robustness.
This approach prevents reduced trajectory diversity by avoiding sole reliance on the maximum-robustness IK solution.
The Optimizer used multithreading with eight threads, and ${T}_{s}$ set to 0.02 s.
For the Evaluator, only the motion duration was considered, disregarding soft constraints.
The $TO_{val}$ for the Validator was set to 0.15 s, with ${VT}_{from\_start}$ at 2 s and $VT_{interval}$ at 0.1 s.
SAMPLE\_ENV returned the point cloud of the evaluation input, independent of the timestamp.

AIT* was implemented using OMPL \cite{ompl}.
CBiRRT2 was executed until a trajectory was found, with a two-path round-robin shortcut for post-processing.
Time optimization was then performed using the optimizer described earlier.

Examples of the trajectories and environments are shown in Fig. \ref{fig:trajectory}.
The evaluation results are presented in Table \ref{tbl:result_with_behavior}.
The evaluation metrics are defined as follows:
\begin{itemize}
    \item Motion Completion Time: The average time required to complete the motion, including both the Plan to Motion Delay and the Motion Duration.
    \item Plan to Motion Delay: The average delay before the motion starts due to trajectory planning.
    If planning fails, the value from CBiRRT2 is used for calculation.
    This assumes a parallel approach where fast planning runs alongside slower, guaranteed planning; if the slower planning finishes first, its result is used.
    \item Motion Duration: The average time the robot is actively moving.
    \item Robustness: The average robustness of the robot state at the end of the motion, as defined by Eq. \ref{eq:robustness_con}.
    \item Collision in Motion Rate: The percentage of instances where a collision occurs during motion.
\end{itemize}

In all methods, including the proposed RLP, the Motion Completion Rate exceeded 99\%, demonstrating the feasibility with minimal waypoints in home environments.
If RLP fails to find a solution in complex cases, RRT is executed as an alternative to generate a more complex trajectory.
While this approach requires more time than executing RRT from the beginning, RLP can still solve 99\% of tasks under diverse conditions, such as those in the BEHAVIOR dataset, enabling the construction of a superior motion planning system.

The RLP exhibited superior performance in both Motion Completion Time and Robustness across all robots.
The RLP had a shorter Motion Duration, as shown in a later ablation study, due to its ability to periodically switch to shorter trajectories.
In terms of robustness, the RLP performed better, particularly with the HSR, which had the minimum DOF.
However, for robots with redundant DOF, such as the Franka MM, robustness remained high across all methods due to the increased likelihood of finding alternative robot states to compensate for errors.
Collisions during motion were minimal for all methods since they were all sampling-based. 
In summary, the RLP enhances computational speed, motion optimality, and robustness in home environments while ensuring safety, regardless of the robot model.
\begin{table}[ht]
    \caption{Performance evaluation results.}
    \label{tbl:result_with_behavior}
    \centering
    \begin{tabular}{l|ccc}
    Robot (Data count)               & \multicolumn{3}{c}{HSR (1488)} \\ \hline
    Method                           & RLP            & AIT* & CBiRRT2 \\
    Periodicity                      & Yes            & No   & No      \\ \hline
    Motion Completion Time{[}sec{]}  & \textbf{5.01}  & 5.28 & 7.22    \\
    \quad\scriptsize{Plan to Motion Delay{[}sec{]}} & \hspace{1mm}\scriptsize{0.46} & \hspace{1mm}\scriptsize{0.45} & \hspace{1mm}\scriptsize{2.32} \\
    \quad\scriptsize{Motion Duration{[}sec{]}}      & \hspace{1mm}\scriptsize{4.55} & \hspace{1mm}\scriptsize{4.83} & \hspace{1mm}\scriptsize{4.90} \\ \hline
    Robustness                       & \textbf{0.73}  & 0.50 & 0.37    \\
    Collision in Motion Rate{[}\%{]} & 0.1            & 0.2  & 0.1      \\
    \end{tabular}

    \vspace{10pt}

    \begin{tabular}{l|ccc}
    Robot (Data count)               & \multicolumn{3}{c}{Franka MM (1348)} \\ \hline
    Method                           & RLP             & AIT*   & CBiRRT2   \\
    Periodicity                      & Yes             & No     & No        \\ \hline
    Motion Completion Time{[}sec{]}  & \textbf{5.03}   & 6.02   & 7.25      \\
    \quad\scriptsize{Plan to Motion Delay{[}sec{]}} & \hspace{1mm}\scriptsize{0.51} & \hspace{1mm}\scriptsize{0.78} & \hspace{1mm}\scriptsize{1.98} \\
    \quad\scriptsize{Motion Duration{[}sec{]}}      & \hspace{1mm}\scriptsize{4.52} & \hspace{1mm}\scriptsize{5.24} & \hspace{1mm}\scriptsize{5.27} \\ \hline
    Robustness                       & \textbf{0.88}   & 0.83   & 0.78      \\
    Collision in Motion Rate{[}\%{]} & 0.4             & 1.0    & 0.4      
    \end{tabular}

    \vspace{-8pt}
\end{table}

\subsection{Ablation Study}
We verified the effects of periodic motion planning and robust IK solutions in the proposed RLP.

Comparisons were made between the proposed RLP and two variations: RLP without SAMPLE\_WITH\_ROBUST\_IK (Algorithm \ref{alg:generator_with_robust_ik}) (RLP-) and a version without periodic updates (RLP\texttt{--}).
The robot models and input sets remain consistent with those outlined in the previous subsection.

The evaluation results are in Table \ref{tbl:result_ablation}.
RLP- exhibited a slightly shorter Motion Completion Time because robustness was not considered. 
This is likely due to the increased computation time for robust IK solutions, which tend to result in larger motions to maintain a safe distance from obstacles, as illustrated in Fig. \ref{fig:trajectory} D and E.
RLP\texttt{--} exhibited a significantly longer Motion Duration, indicating that periodic planning improves motion optimality. 
Overall, robustness was highest for RLP, which explicitly considered it.
\begin{table}[th]
    \vspace{6pt}
    \caption{Ablation study of RLP.}
    \label{tbl:result_ablation}
    \centering
    \begin{tabular}{l|ccc}
    Robot (Data count)               & \multicolumn{3}{c}{HSR (1488)}         \\ \hline
    Method                           & RLP            & RLP- & RLP\texttt{--} \\
    Periodicity                      & Yes            & Yes  & No             \\ \hline
    Motion Completion Time{[}sec{]}  & 5.01  & \textbf{4.94} & 6.34           \\
    \quad\scriptsize{Plan to Motion Delay{[}sec{]}} & \hspace{1mm}\scriptsize{0.46} & \hspace{1mm}\scriptsize{0.44} & \hspace{1mm}\scriptsize{0.43} \\
    \quad\scriptsize{Motion Duration{[}sec{]}}      & \hspace{1mm}\scriptsize{4.55} & \hspace{1mm}\scriptsize{4.50} & \hspace{1mm}\scriptsize{5.91} \\ \hline
    Robustness                       & \textbf{0.73}  & 0.54 & 0.51    \\
    Collision in Motion Rate{[}\%{]} & 0.1            & 0.0  & 0.0      
    \end{tabular}
    
    \vspace{10pt}

    \begin{tabular}{l|ccc}
    Robot (Data count)               & \multicolumn{3}{c}{Franka MM (1348)}   \\ \hline
    Method                           & RLP            & RLP- & RLP\texttt{--} \\
    Periodicity                      & Yes            & Yes  & No             \\ \hline
    Motion Completion Time{[}sec{]}  & 5.03 & \textbf{4.99}  & 6.42      \\
    \quad\scriptsize{Plan to Motion Delay{[}sec{]}} & \hspace{1mm}\scriptsize{0.51} & \hspace{1mm}\scriptsize{0.50} & \hspace{1mm}\scriptsize{0.47} \\
    \quad\scriptsize{Motion Duration{[}sec{]}}      & \hspace{1mm}\scriptsize{4.52} & \hspace{1mm}\scriptsize{4.49} & \hspace{1mm}\scriptsize{5.95} \\ \hline
    Robustness                       & \textbf{0.88}   & 0.79   & 0.77      \\
    Collision in Motion Rate{[}\%{]} & 0.4             & 0.2    & 0.1      
    \end{tabular}

    \vspace{-8pt}
\end{table}

\subsection{Comparison with Learning-Based Methods}
We compared the proposed method with the learning-based method, M$\pi$Nets \cite{mpinets}.

The Franka MM robot model was used in tabletop scenarios, where 1.09 million trajectories were collected as training data for M$\pi$Nets.
In the same environment, 1000 validation data were generated and used for evaluation.

The evaluation results are in Table \ref{tbl:result_mpinets}.
Although RLP had a longer delay time before the motion start, it outperformed the others in terms of motion completion time and other metrics.
The Motion Completion Rate of M$\pi$Nets decreased significantly to 56.9\%.
This decrease is attributed to difficulties in converging to the desired end-effector pose with a mobile manipulator, resulting in a lower completion rate.
\begin{table}[ht]
    \caption{Performance comparison results with M$\pi$Nets.}
    \label{tbl:result_mpinets}
    \centering
    \begin{tabular}{l|cc}
    Method                           & RLP            & M$\pi$Nets \\
    Periodicity                      & Yes            & Yes        \\ \hline
    Motion Completion Time{[}sec{]}  & \textbf{4.20}  & 7.25       \\
    \quad\scriptsize{Plan to Motion Delay{[}sec{]}} & \hspace{1mm}\scriptsize{0.36} & \hspace{1mm}\scriptsize{0.01} \\
    \quad\scriptsize{Motion Duration{[}sec{]}}      & \hspace{1mm}\scriptsize{3.84} & \hspace{1mm}\scriptsize{7.24} \\ \hline
    Robustness                       & \textbf{0.90}   & 0.87      \\
    Collision in Motion Rate{[}\%{]} & 0.1             & 1.1       \\
    Motion Completion Rate {[}\%{]}  & 99.5            & 56.9     
    \end{tabular}

    \vspace{-8pt}
\end{table}

\subsection{Integration Experiment with Tidy-up Task}
The proposed RLP was integrated into a tidy-up task to verify its effectiveness.

\begin{figure}[ht]
    \vspace{6pt}
    \centering
    \includegraphics[width=75mm]{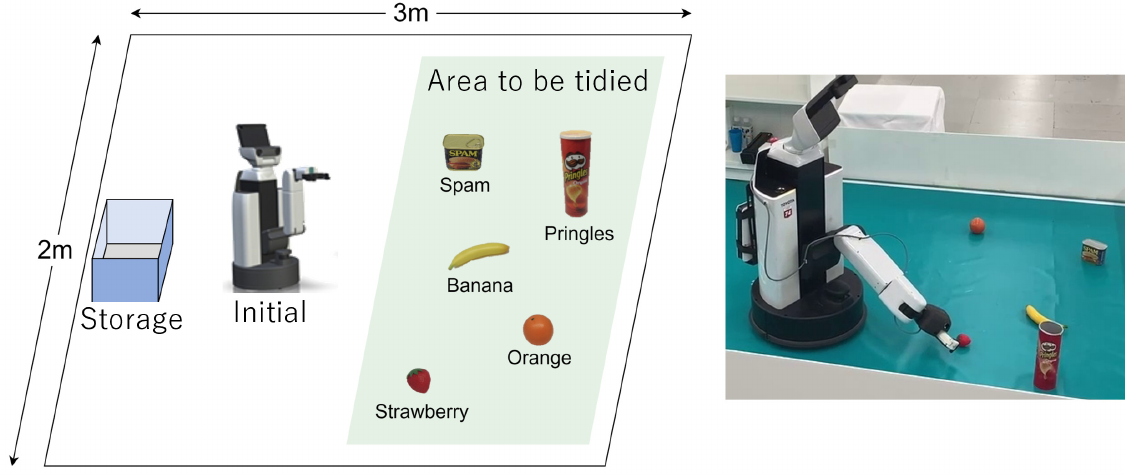}
    \caption{Setup and execution of the tidy-up task.}
	\vspace{-14pt}
    \label{fig:tidyup}
\end{figure}

This simplified tidy-up task, based on the World Robot Summit (WRS) \cite{wrs} (Fig. \ref{fig:tidyup}), involved placing five randomly distributed objects into storage within a specified area, repeated 30 times.
The evaluation metrics included tidy-up time, grasping success rate, and placing success rate.

In the experiments, an HSR was used, equipped with an Orbbec Gemini2 instead of the standard end-effector camera.
For recognition, we used YOLOv8 \cite{yolov8}, XMem++ \cite{xmem2}, and scale-balanced grasp \cite{scale_balanced_grasp} to create a system capable of object recognition, tracking, and grasp pose detection during motion.
Additionally, we implemented a task system that commanded the grasp poses, storage poses, and recognized obstacles for the RLP while controlling the gripper.

The experimental results are presented in Table \ref{tbl:result_pick_and_place}.
Compared with the WRS 2020 winning team system \cite{wrs_ono}, which used CBiRRT2 as the standard motion planner for HSR, our method showed superior performance, particularly in terms of operation time.
This indicates that the RLP’s reduction in motion planning time and motion duration was effective, even in complex manipulation tasks.
\begin{table}[ht]
    \caption{Real-world evaluation in the tidy-up task.}
    \label{tbl:result_pick_and_place}
    \centering
    \begin{tabular}{l|rr}
                                       & \multicolumn{1}{l}{Ours} & \multicolumn{1}{l}{Ono et.al\cite{wrs_ono}} \\ \hline
    Operation time {[}sec{]}           & \textbf{87.6}                  & 122.0                                                \\
    Grasping success rate {[}\%{]}  & \textbf{96.0}                  & 94.0                                                 \\
    Placing success rate {[}\%{]}   & 99.3                     & 99.3                                                
    \end{tabular}

    \vspace{-8pt}
\end{table}

\section{CONCLUSIONS}
This study proposes a periodic sampling-based whole-body trajectory planning method called RLP, designed for rapid and safe manipulation tasks in home environments.
The RLP enables fast planning with minimal waypoints by leveraging the specific characteristics of home environments. 
It optimizes motion through periodic trajectory planning, improves robustness by incorporating IK solutions to handle base position errors, and ensures safety through a sampling-based approach.

Evaluation experiments confirmed that RLP enables efficient motion planning with minimal waypoints in home environments, thereby reducing computational time. 
The method enhances motion optimality and robustness while maintaining safety across different robot models.
In real-world tidy-up tasks, RLP outperformed existing systems in both operation time and success rate, demonstrating its practical effectiveness.


Overall, the RLP enhances computational speed, motion optimality, and robustness while ensuring safety in home environments.
A future challenge will be to verify this method in a dynamic environment.
Given that the RLP performs periodic motion planning, it is expected to adapt to dynamic changes.
However, this capability has not yet been fully verified.

\addtolength{\textheight}{-7cm}   





\newpage


\bibliographystyle{IEEEtran}
\bibliography{reference}

\end{document}